\pdfoutput=1
\documentclass[conference]{IEEEtran}
\usepackage{color}
\usepackage{array}
\usepackage{textcomp}
\usepackage{multirow}
\usepackage{amsmath}
\usepackage{amssymb}
\usepackage{graphicx}
\usepackage{geometry}

\makeatletter

\providecommand{\tabularnewline}{\\}

\IEEEoverridecommandlockouts
\usepackage{cite}
\usepackage{amsfonts}\usepackage{algorithmic}
\usepackage{xcolor}
\def\BibTeX{{\rm B\kern-.05em{\sc i\kern-.025em b}\kern-.08em
   T\kern-.1667em\lower.7ex\hbox{E}\kern-.125emX}}

\makeatother
\newgeometry{left=19.1mm,right=19.1mm,top=25.4mm,bottom=19.1mm}
\begin{document}
\title{A Nested U-Structure for Instrument Segmentation in Robotic Surgery
\thanks{This work was supported in part by the National Natural Science Foundation of China under Grant U2013601 and Grant 62173314 and CAAI-Huawei MindSpore Open Fund.}
\thanks{Y. Xia, S. Wang, and Z. Kan (Corresponding Author) are with the Department
of Automation at the University of Science and Technology of China,
Hefei, Anhui, China, 230026.}}
\author{Yanjie Xia, Shaochen Wang, and Zhen Kan}
\maketitle

\begin{abstract}
Robot-assisted surgery has made great progress with the development
of medical imaging and robotics technology. Medical scene understanding
can greatly improve surgical performance while the semantic segmentation
of the robotic instrument is a key enabling technology for robot-assisted
surgery. However, how to locate an instrument's position and estimate
their pose in complex surgical environments is still a challenging
fundamental problem. In this paper, pixel-wise instrument segmentation
is investigated. The contributions of the paper are twofold: 1) We
proposed a two-level nested U-structure model, which is an encoder-decoder
architecture with skip-connections and each layer of the network structure
adopts a U-structure instead of a simple superposition of convolutional
layers. The model can capture more context information from multiple
scales and better fuse the local and global information to achieve
high-quality segmentation. 2) Experiments have been conducted to qualitatively
and quantitatively show the performance of our approach on three segmentation
tasks: the binary segmentation, the parts segmentation, and the type
segmentation, respectively. The results show that our method significantly
improves the segmentation performance and outperforms state-of-the-art
approaches. 
\end{abstract}

\section{Introduction}

Robot-assisted systems have revolutionized the minimally invasive
surgery to achieve safer, more precise and consistent, and less invasive
intervention \cite{Singh2013}. For instance, the Da Vinci Xi robot
is able to control laparoscopic surgery through remotely operated
by surgeons \cite{BurgnerKahrs2015}. Since the success of robot-assisted
surgery highly relies on the understanding of surgical scene, accurate
segmentation of surgical instruments is crucial.

Recent advances of robotics\cite{Wang2022}\cite{li2021asymmetric} and computer vision technologies
promote the intelligent endoscopic vision, which can help surgeons
perform precise operation. For instance, the augmented reality (AR)
based on endoscopic video can improve surgeon's visual awareness of
high-risk targets\cite{PuertoSouza2014}. The vision-based endoscopic
navigation method has been applied in sinus surgery\cite{Leonard2018}.
The work of \cite{Jin2018} presents a method for automatically assessing
a surgeon's performance by tracking and analyzing tool movements in
surgical videos. The 3D dense reconstruction of handheld monocular
endoscopic surgery scenes was developed in \cite{Mahmoud2018}. However,
in the above applications, the endoscope visual perception is inseparable
from medical scene understanding, resulting in the poor performance
in extracting necessary visual and regional information for surgical
procedures.

Medical scene understanding can significantly improve the surgical
performance, since it can expand the surgeon's perception by providing
information on internal anatomy and surgical instruments. Such information
is usually provided by videos or 2D images consisting of human tissues
and surgical instruments that present their position, shape, size
and posture intuitively. The location of the instrument's position
relative to the patient's anatomy helps surgeons understand surgical
scene and operate more accurately. The pose of the instrument can
be used to measure the distance to risk structures, evaluate the surgeon's
skills, and realize automated surgical operation \cite{Bodenstedt2018}.
Therefore, it is important to extract these valuable information selectively
and intelligently, while avoiding unnecessary information that might
confuse the surgeons.

To address this challenge, scene segmentation of surgical instruments
is a recent research focus, which can separate the instruments from
the background tissue and provides important information in surgical
procedures for surgeons. Segmentation masks can prevent the covering
occlusion apparatus of the rendered tissue and clearly show the position
and pose of the surgical instruments in the endoscopic images\cite{Jin2019}.
Besides, segmentation masks play an important role in instrument tracking
systems. Therefore, the semantic segmentation of surgical robotic
instruments is highly desired for promoting the cognitive assistance
to surgeons. However, due to the complicated medical scene, how to
locate an instrument's position and estimate their pose to achieve
precise segmentation is an essentially fundamental yet challenging
problem\cite{Shvets2018}.

Recently, a variety of vision based methods are developed for the
location and tracking of the instruments\cite{Wang2022a}. Prior methods
of instrument-background segmentation utilized color and texture features\cite{Speidel2006}\cite{Doignon2007},
Haar wavelets\cite{Sznitman2012}, and HoG\cite{Rieke2016}. Later,
machine learning algorithms, such as Random Forest\cite{Bouget2015}
and Gaussian Mixture Model\cite{Pezzementi2009}, were applied to
deal with the segmentation problem. However, these models only focus
on single binary segmentation problems. More complex segmentation,
such as the detection of various parts and types of the instrument,
are desired in modern surgery.

To solve this problem, many deep learning based approaches have been
developed, showing promising performance in medical areas, especially
for tracking, classification
\newgeometry{left=19.1mm,right=19.1mm,top=19.1mm,bottom=19.1mm}
\noindent
and location problems of robotic
surgical instrument. Convolutional neural networks (CNN) have been
successfully applied, which can realize pixel-level segmentation of
images captured by endoscope camera \cite{Long2015}. However, it
requires a large size of training data, which limits their success
in practice. U-Net\cite{Ronneberger2015} with an encoder-decoder
network architecture was designed to address this issue and has achieved
good performance on different biomedical segmentation applications.
In fact, location information is the basis of semantic segmentation.
Accurate location information can lead to precise segmentation performance.
Deep neural networks (DNN) have been used to combine semantic segmentation
with landmark locations \cite{Laina2017}, which learn better feature
representation of the existing input by the training mechanism of
layer by layer via data pre-training. In\cite{Attia2017}, the recurrent
neural network was embedded with convolutional neural network to establish
dependencies among multiple tags. To further improve segmentation
accuracy, \cite{Qin2019} fused the information of kinematic pose
and convolutional neural networks prediction. Besides, \cite{Chaurasia2017}
\cite{Iglovikov2018} \cite{Shvets2018} have provided solutions for
three sub-problems of instrument segmentation, i.e., binary segmentation,
partial segmentation, and type segmentation. While these deep learning-based
methods have achieved impressive results, it still leaves room for
improvement. Moreover, how to improve the accuracy of surgical instrument
segmentation efficiently and make it suitable for multiple segmentation
tasks is still a challenge.

In this paper, to facilitate intelligent surgery, we develop a novel
nested U-structure framework for surgical instrument semantic segmentation.
The goal is to better understand the medical scene and extract the
semantic information to promote minimally invasive surgery. The main
contributions of this work can be summarized as follows: 1) we take
multi-scale feature extraction and multi-level deep feature integration
into consideration and proposed a two-level nested U-structure, which
fuses the local and global features to realize a more precise segmentation
for surgical instrument segmentation tasks. 2) Dilated convolutions
were used in our network to maintain high resolution feature maps
while increasing the reception field of convolution kernel. 3) Experiments
are conducted on the MICCAI EndoVis Challenge 2017 dataset. The results
show that our model can greatly improve the performance of segmentation
and outperforms other state-of-the-art approaches.

\section{METHODS}

\subsection{Overview}

\label{AA} The understanding of medical scene can improve the surgeon's
ability in perception. To have a better scene understanding of the
surgical scenario, semantic segmentation is one of the important methods
to extract the posture and position information of surgical instruments,
which is crucial for the smooth surgical operation. It is especially
helpful for medical imaging and robot-assisted surgical system. Given
an image captured by the high resolution stereo camera, the goal is
to separate the surgical instrument from the background in the image,
and segment the parts and types of surgical instruments semantically.
To achieve this objective, we proposed a new deep learning-based solution.
The overall architecture of our network is illustrated in Fig. 1,
which is a two-level nested U-structure. The images obtained by the
laparoscopic system are taken as input to the network, and the output
is the semantic segmentation of surgical instruments.

\subsection{Network Architecture}

For robot-assisted minimally invasive surgery, when a surgical instrument
is moved and operated within the tissue, the robot needs to locate
and track the instrument. Due to the complex surgical environment,
it is crucial for the model to acquire high-precision segmentation
of instruments in the surgical scene. In general, features from multiple
deep layers are able to generate better results. Taking into account
the memory and the computation budget, we choose a 6-layer deep structure
for our network architecture. For semantic segmentation, both local
and global contextual information are essential for high precision
segmentation. Traditional UNet\cite{Ronneberger2015} uses two $3\times3$
convolutions, rectified linear unit (ReLU), and $2\times2$ maxpooling
operation repeatedly to shrink or expand the feature maps, so as to
extract important feature information. However, more detailed feature
information is required to achieve accurate segmentation and feature
information extraction is limited for simple superposition of convolutional
layers. The new modules need to be designed to implement multi-scale
feature extraction. Inspired by UNet, the model that we proposed adopts
an encoder-decoder architecture which can capture context information
by a contraction path and locate accurately by an expansion path.
Hence, considering multi-level deep feature integration and multi-scale
feature extraction, the network we proposed is a two-level nested
U-structure as shown in Fig. 1(a). Each layer of the network structure
uses a U-structure instead of a simple superposition of convolutional
layers. 

\begin{figure*}[!t]
\begin{center}
\end{center}
\includegraphics[scale=0.64]{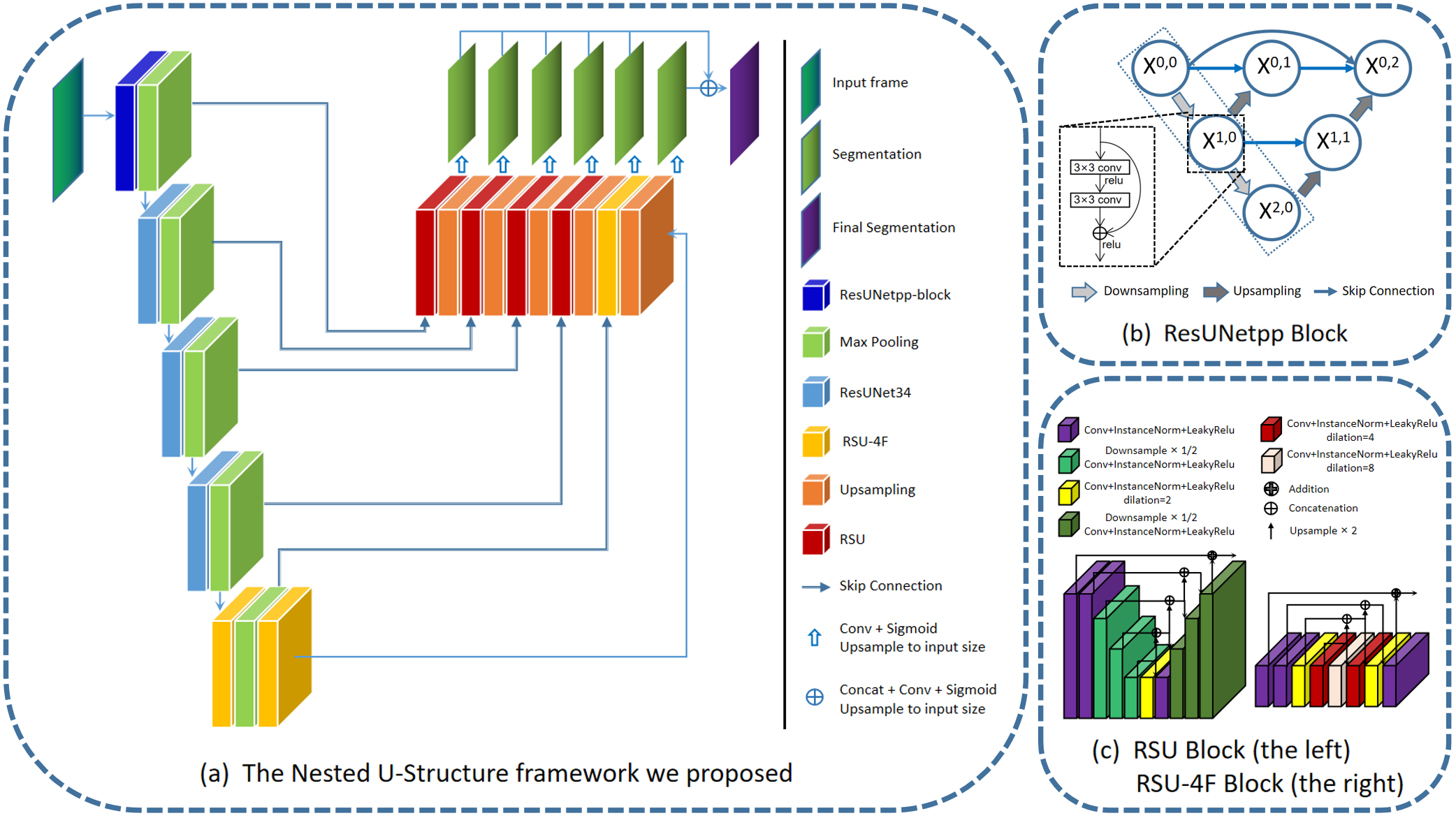}

\caption{The illustration of the proposed network for the semantic segmentation
of surgical instruments.}
\end{figure*}

It is well known that the encode process is essentially a process
of feature extraction while reducing the spatial size of feature maps.
Since modern CNN models have been successfully used and UNet\cite{Ronneberger2015}
has greatly promoted the development of deep learning in the field
of medical imaging, we introduce Resnet into UNet and call it ResUNet.
We use ResUNet as the main encoder backbone of the network architecture.
The starting unit of encoder is a combination of Resnet18 and UNetplusplus\cite{Zhou2019}
which we call it ResUNetpp (Fig. 1(b)). It can be interpreted as UNetplusplus
using Resnet18 as the encoder. Subsequently, there are 4 stages of
encoder which consists of ResUNet34. Similarly, ResUNet34 can be interpreted
as UNet using Resnet34 as the encoder. Each step in the contraction
path contains alternating convolution and pooling operations, and
the feature maps are gradually downsampled with stride 2 while increasing
the number of feature maps at each layer. After 4 times of downsampling,
the resolution of the feature map has been greatly reduced. The feature
information will be lost if we continue on downsampling. Therefore,
we adopt the RSU-4F\cite{Qin2020} module at the bottom of the structure
to keep the resolution consistent between the input and output feature
maps. As shown in Fig. 1(c), RSU-4F is a four layer structure similar
to UNet which consists of a $3\times3$ convolutions, a Batch Normalization(BN)
and a rectified linear unit (ReLU). To maintain high resolution feature
maps while increasing the reception field of convolution kernel, we
use dilated convolutions to upsample and downsample. The expanding
path increases the resolution of the feature maps by upsampling and
the backbone of decoder also follows a similar architecture of UNet.
The decoder consists of 4 ResSdual Ublocks(RSU)\cite{Qin2020} and
the architecture of RSU is shown in Fig. 1(c). It is a variant of
a 5-layer UNet which enables the network architecture to extract the
features of multiple scales from each residual block. Skip-connections
have been applied to combine the feature maps from contracting path
and expanding path. It is worth mentioning that we have replaced the
classical activation function ReLU with LeakyReLU and substitute InstanceNorm2d
for BatchNorm2d in our model. The network takes RGB images as input
and generate pixel-level segmentation prediction pictures. We perform
three segmentation tasks by setting the number of output channels
of the network structure. 

\begin{figure*}[!t]
\begin{center}
\end{center}
\includegraphics[scale=0.758]{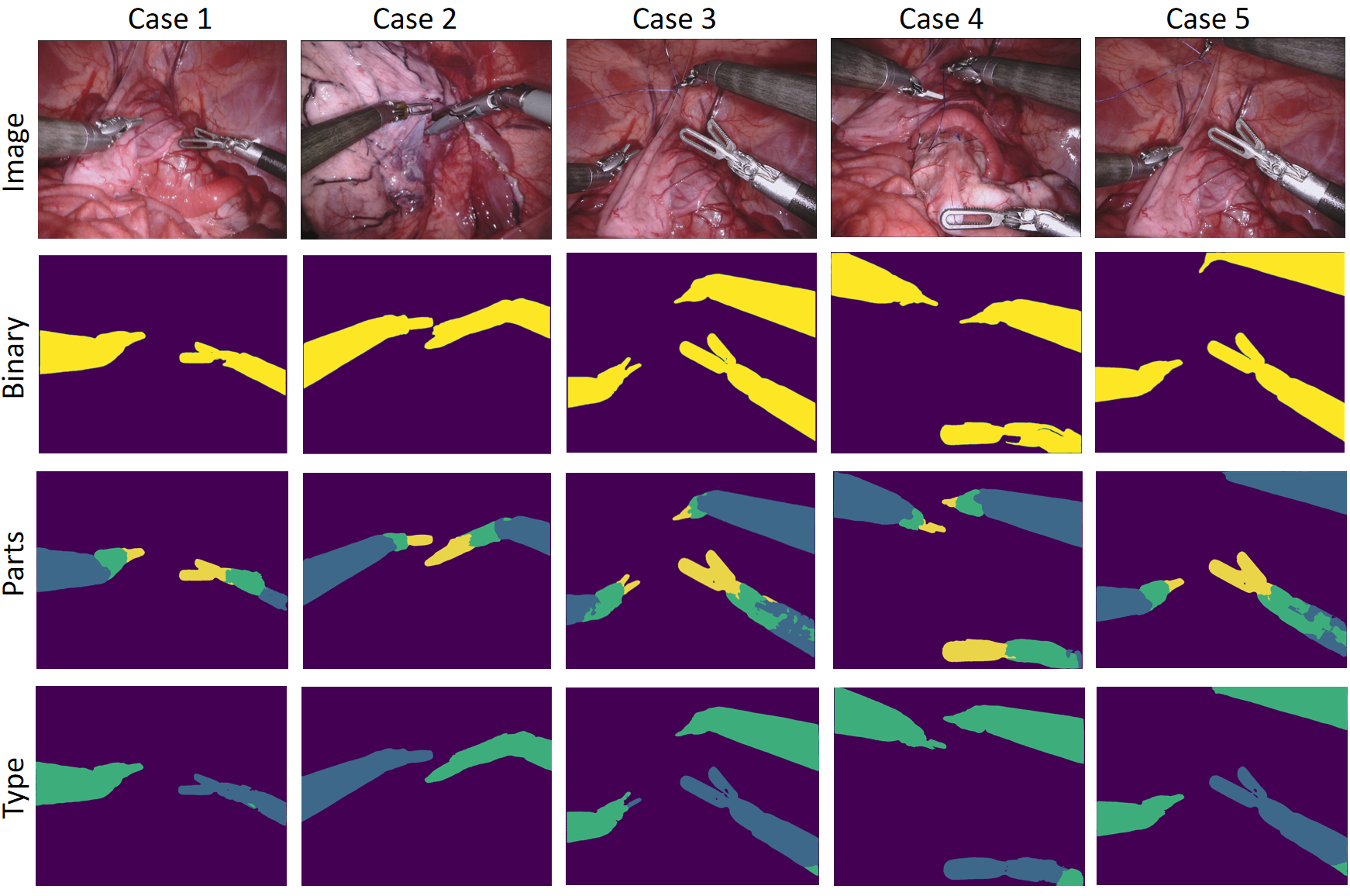}

\caption{Examples of visual segmentation results of the proposed model.}
\end{figure*}

\subsection{Lossfunction}

Since image segmentation tasks can be regarded as a classification
problem of pixels, the overlap rate between the predicted masks and
the corresponding ground truth represents the probability that the
pixel belongs to each category. The segmentation loss function is
based on Jaccard index, which indicates the similarity between two
sets. In this work, we introduce a common loss function, denoted as
$H$. In different segmentation tasks, $H$ represents different loss
functions. For binary segmentation task, $H$ adopts the BCEWithLogitsLoss.
For multi-class segmentation problem, $H$ represents the\textcolor{blue}{{}
}Cross-Entropy Loss. We define the generalized segmentation loss function
as follows: 
\begin{equation}
Loss=H-log(\frac{1}{n}\sum_{i=1}^{n}(\frac{m_{i}n_{i}}{m_{i}{_{+}}n_{i}{_{-}}m_{i}n_{i}})),
\end{equation}
where $m_{i}$ and $n_{i}$ represent the ground truth and the predicted
output for the pixel $i$, respectively. In order to perform the segmentation
task better, it has to minimize the generalized segmentation loss
function via maximizing the probability of correctly predicting pixels.

\begin{table*}
\caption{Comparison of instrument segmentation results on the three tasks(mean\textpm std).}

\noindent \centering{}%
\begin{tabular}{|>{\centering}p{2cm}|>{\centering}p{2.1cm}>{\centering}p{2.1cm}|>{\centering}p{2.1cm}>{\centering}p{2.1cm}|>{\centering}p{2.1cm}>{\centering}p{2.1cm}|}
\hline 
\multirow{1}{2cm}{Methods} & \multicolumn{2}{c|}{Binary segmentation} & \multicolumn{2}{c|}{Parts segmentation} & \multicolumn{2}{c|}{Type segmentation}\tabularnewline
\hline 
 & IOU(\%)  & Dice(\%)  & IOU(\%)  & Dice(\%)  & IOU(\%)  & Dice(\%)\tabularnewline
\hline 
U-Net  & 75.44 \textpm{} 18.18  & 84.37 \textpm{} 14.58  & 48.41 \textpm{} 17.59  & 60.75 \textpm{} 18.21  & 15.80 \textpm{} 15.06  & 23.59 \textpm{} 19.87\tabularnewline
\hline 
TernausNet  & 81.14 \textpm{} 19.11  & 88.07 \textpm{} 14.63  & 62.23 \textpm{} 16.48  & 74.25 \textpm{} 15.55  & 34.61 \textpm{} 20.53  & 45.86 \textpm{} 23.20\tabularnewline
\hline 
LinkNet-34  & 82.36 \textpm{} 18.77  & 88.87 \textpm{} 14.35  & 34.55 \textpm{} 20.96  & 41.26 \textpm{} 23.44  & 22.47 \textpm{} 35.73  & 24.71 \textpm{} 37.54\tabularnewline
\hline 
PlainNet  & 81.86 \textpm{} 15.85  & 88.96 \textpm{} 12.98  & \textbf{64.73 \textpm{} 17.39}  & \textbf{73.53 \textpm{} 16.98}  & 34.57 \textpm{} 21.93  & 44.64 \textpm{} 25.16\tabularnewline
\hline 
Ours  & \textbf{82.94 \textpm{} 16.82}  & \textbf{89.42 \textpm{} 14.01}  & 58.38 \textpm{} 19.06  & 69.59 \textpm{} 18.66  & \textbf{41.72 \textpm{} 33.44}  & \textbf{48.22 \textpm{} 34.46}\tabularnewline
\hline 
\end{tabular}
\end{table*}

\begin{table*}
\caption{The numerical results of our method and comparison with other methods
in binary segmrntation of robotic tools.}

\centering{}%
\begin{tabular}{>{\centering}p{1cm}|>{\centering}p{1cm}|>{\centering}p{1cm}|>{\centering}p{1cm}|>{\centering}p{1cm}|>{\centering}p{1cm}|>{\centering}p{1cm}|>{\centering}p{1cm}|>{\centering}p{1cm}|>{\centering}p{1cm}|>{\centering}p{1cm}|>{\centering}p{1cm}}
\hline 
\noindent \centering{}  & \noindent \centering{}Dataset 1  & \noindent \centering{}Dataset 2  & \noindent \centering{}Dataset 3  & \noindent \centering{}Dataset 4  & \noindent \centering{}Dataset 5  & \noindent \centering{}Dataset 6  & \noindent \centering{}Dataset 7  & \noindent \centering{}Dataset 8  & \noindent \centering{}Dataset 9  & \noindent \centering{}Dataset 10  & \noindent \centering{}mIOU\tabularnewline
\hline 
\noindent \centering{}NCT  & \noindent \centering{}0.784  & \noindent \centering{}0.788  & \noindent \centering{}0.926  & \noindent \centering{}0.934  & \noindent \centering{}0.701  & \noindent \centering{}0.876  & \noindent \centering{}0.846  & \noindent \centering{}0.881  & \noindent \centering{}0.789  & \noindent \centering{}0.899  & \noindent \centering{}0.843\tabularnewline
\noindent \centering{}UB  & \noindent \centering{}0.807`  & \noindent \centering{}0.806  & \noindent \centering{}0.914  & \noindent \centering{}0.925  & \noindent \centering{}0.740  & \noindent \centering{}0.890  & \noindent \centering{}\textbf{0.930}  & \noindent \centering{}0.904  & \noindent \centering{}0.855  & \noindent \centering{}\textbf{0.917}  & \noindent \centering{}0.875\tabularnewline
\noindent \centering{}BIT  & \noindent \centering{}0.275  & \noindent \centering{}0.282  & \noindent \centering{}0.455  & \noindent \centering{}0.310  & \noindent \centering{}0.220  & \noindent \centering{}0.338  & \noindent \centering{}0.404  & \noindent \centering{}0.366  & \noindent \centering{}0.236  & \noindent \centering{}0.403  & \noindent \centering{}0.326\tabularnewline
\noindent \centering{}MIT  & \noindent \centering{}0.854  & \noindent \centering{}0.794  & \noindent \centering{}0.949  & \noindent \centering{}0.949  & \noindent \centering{}\textbf{0.862}  & \noindent \centering{}\textbf{0.922}  & \noindent \centering{}0.856  & \noindent \centering{}0.937  & \noindent \centering{}0.865  & \noindent \centering{}0.905  & \noindent \centering{}0.888\tabularnewline
\noindent \centering{}SIAT  & \noindent \centering{}0.625  & \noindent \centering{}0.669  & \noindent \centering{}0.897  & \noindent \centering{}0.907  & \noindent \centering{}0.604  & \noindent \centering{}0.843  & \noindent \centering{}0.832  & \noindent \centering{}0.513  & \noindent \centering{}0.839  & \noindent \centering{}0.899  & \noindent \centering{}0.803\tabularnewline
\noindent \centering{}UCL  & \noindent \centering{}0.631  & \noindent \centering{}0.645  & \noindent \centering{}0.895  & \noindent \centering{}0.883  & \noindent \centering{}0.719  & \noindent \centering{}0.852  & \noindent \centering{}0.710  & \noindent \centering{}0.517  & \noindent \centering{}0.808  & \noindent \centering{}0.869  & \noindent \centering{}0.785\tabularnewline
\noindent \centering{}TUM  & \noindent \centering{}0.760  & \noindent \centering{}0.799  & \noindent \centering{}0.916  & \noindent \centering{}0.915  & \noindent \centering{}0.810  & \noindent \centering{}0.873  & \noindent \centering{}0.844  & \noindent \centering{}0.895  & \noindent \centering{}\textbf{0.877}  & \noindent \centering{}0.909  & \noindent \centering{}0.873\tabularnewline
\noindent \centering{}Delhi  & \noindent \centering{}0.408  & \noindent \centering{}0.524  & \noindent \centering{}0.743  & \noindent \centering{}0.782  & \noindent \centering{}0.528  & \noindent \centering{}0.292  & \noindent \centering{}0.593  & \noindent \centering{}0.562  & \noindent \centering{}0.626  & \noindent \centering{}0.715  & \noindent \centering{}0.612\tabularnewline
\noindent \centering{}UA  & \noindent \centering{}0.413  & \noindent \centering{}0.463  & \noindent \centering{}0.703  & \noindent \centering{}0.751  & \noindent \centering{}0.375  & \noindent \centering{}0.667  & \noindent \centering{}0.362  & \noindent \centering{}0.797  & \noindent \centering{}0.539  & \noindent \centering{}0.689  & \noindent \centering{}0.591\tabularnewline
\noindent \centering{}UW  & \noindent \centering{}0.337  & \noindent \centering{}0.289  & \noindent \centering{}0.483  & \noindent \centering{}0.678  & \noindent \centering{}0.219  & \noindent \centering{}0.619  & \noindent \centering{}0.325  & \noindent \centering{}0.506  & \noindent \centering{}0.377  & \noindent \centering{}0.603  & \noindent \centering{}0.461\tabularnewline
\hline 
\noindent \centering{}Ours  & \noindent \centering{}\textbf{0.877}  & \noindent \centering{}\textbf{0.814}  & \noindent \centering{}\textbf{0.962}  & \noindent \centering{}\textbf{0.959}  & \noindent \centering{}0.849  & \noindent \centering{}0.892  & \noindent \centering{}0.812  & \noindent \centering{}\textbf{0.956}  & \noindent \centering{}0.855  & \noindent \centering{}\textbf{0.917}  & \noindent \centering{}\textbf{0.889}\tabularnewline
\hline 
\end{tabular}
\end{table*}

\section{EXPERIMENTS}

In this section, experiments are conducted to show the performance
of the proposed network architecture in three types segmentation tasks.
To evaluate the accuracy of this nested U-structure for segmentation
qualitatively and quantitatively, we calculate the Intersection Over
Union (IOU), also referred to Jaccard Index, as the evaluation criteria.
Meanwhile, we use Dice coefficient (Dice) as another evaluation metrics.
The segmentation accuracy is proportional to the numerical value of
IOU and Dice. We compared our approach with the state-of-the-art methods
on EndoVis 2017 dataset, and analyzed the experimental results.

\subsection{Dataset}

The dataset we used in this paper is provided by the Endoscopic Vision
Challenge 2017\cite{Allan2019}. The training dataset consists of
8 robotic surgical videos acquired from da Vinci Xi surgical system
in different procedures, and each video is divided into a sequence
of 225 images. To avoid data redundancy, video sampling rate of 2
Hz in the training sequences was provided. The RGB stereo channels
from the left and right cameras together form these video sequences.
The images taken by the camera on the left provide the hand-labeled
ground truth for every robotic instrument, but the right frames was
not provided. Therefore, the training images are from the left channel.
Different surgical instruments such as rigid shafts, articulated wrists,
claspers, drop-in ultrasound probe, and a laparoscopic instrument,
have all been labelled by hand. A surgical instrument can be roughly
divided into three parts: shaft, wrist, and clasper which are also
labelled in the frames.

The testing dataset consists of $8\times75$ frame sequences sampled
immediately after each training sequence and two full 300-frame sequences.
These sequences were sampled at the same rate as the training set,
resulting in ten test datasets with a total of 1200 images.

\subsection{Training}

Before starting the training, the input images are pre-processed.
Every RGB images generated from surgical video sequences have a high
resolution of $1920\times1080$ pixels. In order to crop out the black
canvas in the frames, images should be reduced to $1280\times1024$
which are necessary to be cropped at the position of $\left(320,28\right)$.
Besides, several simple augmentations (e.g., PadIfNeeded, RandomCrop,
Flip Horizontal and Flip vertical) are used for dataset's pre-processing
in order to improve the performance of semantic segmentation. The
dataset within each image channel is normalized and the mean value
of each channel is subtracted to get a zero-average image.

To compare directly and fairly, IOU is a standard performance measure
for segmentation problems. The IoU value is calculated as 
\begin{equation}
IOU=\frac{1}{n}\sum_{i=1}^{n}(\frac{m_{i}n_{i}}{m_{i}+n-mn_{i}}),
\end{equation}
where $m_{i}$ and $n_{i}$ represent the ground truth value and the
predicted output for the pixel $i$, respectively.

To measure the similarity of the sets, Dice is defined as : 
\begin{equation}
Dice=\sum_{i=1}^{n}\frac{2\times TP_{i}}{2\times TP_{i}+FP_{i}+FN_{i}},
\end{equation}
where the $n$ represent the number of images in the dataset for evaluation,
and $TP_{i}$, $FP_{i}$, and $FN_{i}$ denote the numbers of true
positives, the false positives and the false negatives for every image,
respectively. The value of Dice is between 0 and 1, which indicates
the similarity of the semantic segmentation.

In experiments, we set the learning rate as 1e-4. During the training,
the optimizer we used in our proposed model is AdamW. In order to
evaluate the generality of the model to the data, the K-fold cross-validation
is employed in the course of training, which divides the dataset into
4 folds. Our framework is implemented in MindSpore with 2 NVIDIA GTX3090
GPUs for training. We trained all the models for 100 epochs with the
batch size set to 2. Multiple GPUs are allowed to accelerate the training.

\subsection{Results}

In order to verify the performance of the network qualitatively and
quantitatively, we conducted a series of segmentation experiments
based on EndoVis 2017 dataset. We performed a qualitative comparison
with state-of-the-art models, such as U-Net\cite{Ronneberger2015},
TernausNet\cite{Iglovikov2018}, LinkNet-34\cite{Shvets2018} and
PlainNet\cite{Jin2019}, and listed the results of three segmentation
tasks in the TABLE I. As we shown in the TABLE I, for binary segmentation,
our method achieves the best result, i.e., the IOU score of 82.94\%
and the Dice score of 89.42\%. In particular, 1) compared to U-Net,
our model achieves an improvement of 7.5 points for IOU and 5.05 points
for Dice. 2) compared to prior advanced methods, we still obtain the
best segmentation performance. For parts segmentation, our method
does not show the best performance compared to TernausNet and LinkNet-34,
but it still greatly improves its segmentation performance over the
UNet for 9.97 points. For the task of multi-class class instrument
segmentation, as we list in the table, our network achieves the best
performance with the IOU score of 41.72\% and the Dice score of 48.22\%
and the result is far superior to others methods. Since several of
the seven categories of instruments only appear a few times in the
training dataset, the performance of this task is overall lower. The
result suggests that increasing the size of the dataset for the corresponding
problem can effectively improve the performance.

We have also demonstrated a more intuitive result by visualizing the
result of the segmentation tasks of our model on the dataset in Fig.
2. There are three different sub-tasks, i.e., binary segmentation
(2 classes), part of instrument segmentation (4 classes), and instrument
type segmentation (8 classes). For binary and parts segmentation,
we encode the ground truth labels with values $(10,20,30,40,0)$ to
distinguish the background and every part of an instrument. Besides,
the instrument type labels are used to classify different surgical
instruments, and they are encoded with an incremental numerical value
starting from 1 to 7. In order to display the segmentation effect
more clearly, we convert the image of the segmentation result into
color. As shown in the figure, the instruments and backgrounds are
distinguished by purple and yellow, respectively. Three parts of each
instrument can be identified individually by different colors while
yellow represents the clasper, green represents the wrist and blue
represents the shaft, respectively. For the type segmentation, the
seven classes of instruments are also distinguished by different colors.
Fig. 2 shows that our model can basically complete the detection and
segmentation of instrument edges and types well. For case 3 and case
5, for parts segmentation, the surgical instrument in the lower right
corner of the picture is not segmented very well. The possible explanation
is that it was caused by the reflection of the light from the instrument.

In addition, we evaluate our trained network for instrumentation segmentation
on ten different test video sequences, which consists of $8\times75$
frame sequences and two full 300-frame sequences. TABLE II lists the
performances of our method and that of other ten teams. We show the
test result of our proposed model in ten datasets and compare it to
the results of ten teams. As shown in the table, it is noticeable
that the method we proposed achieves the highest IOU score in 6 datasets.
This result demonstrates the high efficiency and accuracy of our model
for the semantic segmentation task.

\section{CONCLUSION}

In this work, we present a novel model for robotic surgical instrument
segmentation, which can address three kinds of surgical instrument
segmentation tasks in surgical scenes. The model we proposed is a
nested U-structure which is based on the network architecture of UNet.
Our method is compared with the existing state-of-the-art models in
terms of IOU and Dice, and can achieve efficient and accurate segmentation.
We also present comparative analysis of multiple deep network models
through experimental data. The experimental results suggest that our
model can highly optimize the surgical instrument segmentation and
has achieved highly competitive performance for three sub-tasks, especially
for the binary instrumentation segmentation and the type instrument
segmentation.

\bibliographystyle{IEEEtran}
\bibliography{bib}

\end{document}